\documentclass[runningheads]{llncs}

\usepackage{eccv}

\usepackage{eccvabbrv}

\usepackage{graphicx}
\usepackage{booktabs}

\usepackage{multirow}
\usepackage{tabularx}
\newcolumntype{Y}{>{\centering\arraybackslash}X}
\newcolumntype{C}[1]{>{\centering}p{#1}}
\newcolumntype{Z}{>{\raggedleft\arraybackslash}X}

\usepackage{pifont}
\usepackage{xcolor}
\newcommand{\cmark}{\textcolor{green!80!black}{\ding{51}}}
\newcommand{\xmark}{\textcolor{red}{\ding{55}}}

\usepackage[accsupp]{axessibility}  

\usepackage{hyperref}

\usepackage{orcidlink}
\usepackage{comment}

\newcommand{\ds}{{SpaRe}}
\newcommand{\bench}{{SpaRe}}

\usepackage{amsfonts}
\makeatletter
\newcommand*\tablesmallsize{%
  \@setfontsize\tablesmallsize{8}{9}%
}
\makeatother

\begin{document}

\title{AIM 2024 Sparse Neural Rendering Challenge:\\Dataset and Benchmark} 

\titlerunning{SpaRe Dataset and Benchmark}


\author{Michal Nazarczuk\inst{1} \and Thomas Tanay\inst{1} \and Sibi Catley-Chandar\inst{1} \and\\Richard Shaw\inst{1} \and Radu Timofte\inst{2} \and Eduardo Pérez-Pellitero\inst{1}
}


\authorrunning{M.~Nazarczuk et al.}

\institute{Huawei Noah's Ark Lab, London, United Kingdom\\
\email{\{michal.nazarczuk1, thomas.tanay, sibi.catley.chandar,\\richard.shaw1, e.perez.pellitero\}@huawei.com}
\and University of Würzburg, Germany \\
\email{radu.timofte@uni-wuerzburg.de}
}

\maketitle

\begin{abstract}
Recent developments in differentiable and neural rendering have made impressive breakthroughs in a variety of 2D and 3D tasks, \eg novel view synthesis, 3D reconstruction. Typically, differentiable rendering relies on a dense viewpoint coverage of the scene, such that the geometry can be disambiguated from appearance observations alone. Several challenges arise when only a few input views are available, often referred to as \textit{sparse} or \textit{few-shot} neural rendering. As this is an underconstrained problem, most existing approaches introduce the use of regularisation, together with a diversity of learnt and hand-crafted priors. A recurring problem in sparse rendering literature is the lack of an homogeneous, up-to-date, dataset and evaluation protocol. While high-resolution datasets \cite{barron2022} are standard in dense reconstruction literature, sparse rendering methods often evaluate with low-resolution images. Additionally, data splits are inconsistent across different manuscripts, and testing ground-truth images are often publicly available, which may lead to over-fitting. In this work, we propose the \textbf{Spa}rse \textbf{Re}ndering (\ds) dataset and benchmark. We introduce a new dataset that follows the set-up of the DTU MVS dataset \cite{aanaes2016}. The dataset is composed of 97 new scenes based on synthetic, high-quality assets. Each scene has up to 64 camera views and 7 lighting configurations, rendered at $1600\times1200$ resolution. We release a training split of 82 scenes to foster generalizable approaches, and provide an online evaluation platform for the validation and test sets, whose ground-truth images remain hidden. We propose two different sparse configurations (3 and 9 input images respectively). This provides a powerful and convenient tool for reproducible evaluation, and enable researchers easy access to a public leaderboard with the state-of-the-art performance scores. Available at \url{https://sparebenchmark.github.io}.
\end{abstract}
\section{Introduction}

The recent advancements of neural rendering techniques have catalyzed significant progress in a wide array of computer vision tasks, from novel view synthesis to 3D reconstruction. However, most research in this domain has predominantly focused on dense-view input data, often overlooking the challenges and opportunities posed by sparse data configurations. To address this gap, we introduce \ds{} — the \textbf{Spa}rse \textbf{Re}ndering dataset and benchmark — a novel dataset explicitly designed to benchmark and advance the state-of-the-art in sparse-view neural rendering. \ds{} was developed as the core dataset for the AIM 2024 Sparse Neural Rendering Challenge~\cite{Nazarczuk2024_report}, which aimed to spur innovation in the field of few-shot neural rendering.

Sparse-view rendering presents unique challenges, as it requires robust methods capable of generating high-fidelity reconstructions from a minimal set of input views, often with little overlap between neighbouring images. While existing datasets have primarily catered to dense, well-sampled scenarios, \ds{} intentionally shifts the focus to scenarios where only sparse input data is available, simulating practical conditions in areas like autonomous driving, augmented reality, and resource-constrained environments. 

Current neural rendering datasets face several limitations that impact their effectiveness in training and evaluating models such as Neural Radiance Fields (NeRF)\,\cite{mildenhall2020} for the task of sparse-view rendering. A key limitation is often the quality of the ground-truth data, with many datasets, \eg \cite{mildenhall2019local, barron2022, wizadwongsa2021, wang2023, zhou2018stereo, deitke2023objaverse}, containing potentially imprecise camera poses typically estimated via COLMAP\cite{schoenberger2016sfm}. Inaccurate calibration can lead to errors in model training, which is especially detrimental in the sparse-view setting where precise camera poses are crucial.
The NVS-RGBD dataset~\cite{wang2023} introduced with SparseNeRF, contains real-world scenes with ground-truth depth maps captured using depth sensors, however, the resulting depths are often of lower resolution, inaccurate and noisy, while the camera poses are estimated using COLMAP, impeding reliable evaluation. \ds{} addresses these limitations with synthetically generated data with precise camera poses and accurately rendered high-resolution ground-truth images.

Another limitation with using existing datasets for benchmarking the performance of sparse rendering methods is that they are usually either entirely real~\cite{mildenhall2019local, barron2022} or entirely synthetic~\cite{mildenhall2020, shapenet2015}. Synthetic-only datasets might not generalize well to real-world scenarios, where lighting, reflections, and textures are more complex, whereas, as mentioned previously, real-only datasets often lack precise ground-truth information.
Moreover, the difference between synthetic and real-world datasets can cause models trained on synthetic datasets to perform poorly when applied to real-world scenes. Therefore, the \ds{} benchmark contains both synthetically generated scenes and real-world scenes carefully selected from the DTU dataset~\cite{aanaes2016}. The \ds{} synthetic scenes are specially designed to minimize the domain gap between synthetic and real data by closely reproducing the DTU capture setup (camera poses and light positions).

Currently, the evaluation protocols for sparse-view rendering using the DTU dataset are inconsistent across the literature, often relying on outdated setups, such as evaluating performance at lower resolution. This approach can lead to inflated PSNR scores, as low-resolution novel view synthesis (NVS) is typically easier than high resolution, and may obscure the true computational challenges involved in rendering at higher resolutions. Additionally, different manuscripts use varying DTU versions with different test splits and image interpolation methods, further affecting PSNR. The pixelNeRF~\cite{yu2021} protocol, which uses the same camera views for training and testing for all objects, risks overfitting of generalizable models. To address these issues, the \ds{} online benchmark platform standardizes evaluation by using full-resolution images and a hidden ground-truth test set, ensuring reproducibility and reliability.

The \ds{} dataset consists of 102 meticulously curated synthetic scenes based on high-quality photorealistic 3D assets, each rendered from a sparse set of camera viewpoints for testing (3 or 9 input images), while dense view coverage is provided in the training set. Each scene has up to 64 camera views centred on a single 3D object, with 16 point lights providing 7 different lighting configurations, rendered at $1600\times1200$ resolution. The dataset includes a diverse range of scenes varying in complexity, materials, textures and occlusions, providing a comprehensive testbed for evaluating the performance of sparse neural rendering algorithms, examples of which are shown in Fig.~\ref{fig:examples}. 

This paper details the construction of the \ds{} dataset, including its design principles, data generation processes, evaluation protocol, and the associated challenge tasks. We further discuss the performance of top-performing methods from the AIM 2024 Sparse Neural Rendering Challenge, providing insights into current capabilities and areas ripe for future research in sparse neural rendering. Through \ds{}, we aim to provide the research community with a valuable resource that drives the development of more efficient and effective neural rendering methods in sparse data environments.

\section{Related Work}

\subsection{Datasets}

A number of datasets have been used in the context of sparse novel view synthesis before. We briefly review them here in chronological order.

\begin{itemize}
\item The DTU dataset~\cite{aanaes2016} was originally introduced for multiview stereo evaluation. It consists of 124 scenes/objects placed on a table and photographed 49 or 64 times under 7 different illuminations, using a camera placed on an industrial robot arm encaged in a black box. This dataset has been used extensively for evaluating sparse neural rendering methods \cite{yu2021, chibane2021stereo, mvsnerf21, jain2021, niemeyer2022, kim2022, kangle2022dsnerf, johari2022geonerf, suhail2022generalizable, yang2023, truong2023, wynn2023, seo2023, wang2023, tanay2024, xu2024, li2024, wu2024, chen2024, zhang2024}, following various protocols introduced in PixelNeRF~\cite{yu2021}, RegNeRF~\cite{niemeyer2022} or MVSNeRF~\cite{mvsnerf21}. We discuss the limitations of these protocols in Section~\ref{sec:curr_eval}.
\item The RealEstate10K dataset~\cite{zhou2018stereo} was introduced with a generalisable multiplane image model for stereo magnification. It consists of a large number of camera poses corresponding to 10 million frames derived from about 80,000 video clips, gathered from about 10,000 YouTube videos. Using it requires to extract the corresponding video frames from YouTube. This is a large but not very diverse dataset, and the quality of the images is relatively low. 
\item The Spaces~\cite{flynn2019deepview} was introduced with DeepView, another generalisable multiplane image model. It consists of a 100 indoor and outdoor scenes captured 5 to 10 times each using a 16-camera rig translated by small amounts. 
\item ShapeNet~\cite{shapenet2015} is a large-scale repository of simple shapes represented by 3D CAD models of objects. It has been used for the evaluation of some early sparse neural rendering approaches~\cite{sitzmann2019scene,yu2021}.
\item The LLFF dataset~\cite{mildenhall2019local} was introduced with the generalisable multiplane image model of the same name. It consists of 8 forward facing scenes captured with a phone and camera poses were estimated using COLMAP~\cite{schoenberger2016sfm}. This dataset has also been used extensively for the evaluation of sparse neural rendering approaches~\cite{mvsnerf21,tanay2024,niemeyer2022,seo2023,kangle2022dsnerf,wang2023,wu2024,gao2024cat3d}. 
\item The Blender dataset~\cite{mildenhall2020}, introduced with NeRF~\cite{mildenhall2020}, consists of 8 synthetic scenes viewed from a half-sphere.
\item The IBRNet dataset~\cite{wang2021ibrnet} contains 67 forward-facing scenes captured in a setup similar to LLFF. These additional scenes allow the training of generalisable models.
\item The Shiny dataset~\cite{wizadwongsa2021} was introduced with the NeX multiplane image model. It consists of 8 scenes containing diverse reflective surfaces, and was collected with a focus on evaluating complex non-Lambertian effects. 
\item The Common Objects in 3D dataset (CO3D)~\cite{reizenstein2021common} is a large dataset of real object-centric scenes. It contains a total of 1.5 million frames from nearly 19,000 videos capturing objects from 50 MS-COCO categories.
\item The Mip-NeRF360 dataset~\cite{barron2022} consists of 9 scenes with 5 outdoors and 4 indoors, each containing a complex central object or area with a detailed background.
\item The Objaverse dataset~\cite{deitke2023objaverse} and its follow-up XL version \cite{deitke2024objaverse} is a very large dataset of 10M+ 3D Objects. It has recently enabled rapid progress in the development of large object-centric 3D generative models~\cite{liu2023zero,liu2024one,shimvdream,gao2024cat3d}. 
\end{itemize}

\subsection{Sparse View Rendering}

\begin{table}[p]
\centering
\caption{Overview of sparse neural rendering methods, sorted chronologically.}
\begin{tabularx}{\linewidth}{X|c|c|c}  
    Method & NeRF & 3DGS & Prior \\ 
    \hline
    Stereo Mag.\,\cite{zhou2018stereo} & \xmark & \xmark & Learned (generalisable) \\
    LLFF\,\cite{mildenhall2019local} & \xmark & \xmark & Learned (generalisable) \\ 
    DeepView\,\cite{flynn2019deepview} & \xmark & \xmark & Learned (generalisable) \\ 
    SRN\,\cite{sitzmann2019scene} & \cmark & \xmark & Learned (generalisable) \\
    PixelNeRF\,\cite{yu2021} & \cmark & \xmark & Learned (generalisable) \\ 
    IBRNet\,\cite{wang2021ibrnet} & \xmark & \xmark & Learned (generalisable) \\ 
    SRF\,\cite{chibane2021stereo} & \cmark & \xmark & Learned (generalisable) \\
    MVSNeRF\,\cite{mvsnerf21} & \cmark & \xmark & Learned (generalisable) \\ 
    DietNeRF~\cite{jain2021} & \cmark & \xmark & {CLIP-based semantic consistency loss} \\ 
    RegNeRF~\cite{niemeyer2022} & \cmark & \xmark & {Depth smoothness,
    normalizing flow likelihood} \\ 
    InfoNeRF~\cite{kim2022} & \cmark & \xmark & {Entropy constraint on rays} \\ 
    DDP~\cite{roessle2022} & \cmark & \xmark & {SfM pointcloud with depth completion} \\ 
    DS-NeRF\,\cite{kangle2022dsnerf} & \cmark & \xmark & {SfM pointcloud} \\ 
    GeoNeRF\,\cite{johari2022geonerf} & \cmark & \xmark &  Learned (generalisable) \\ 
    GPNR\,\cite{suhail2022generalizable} & \xmark & \xmark &  Learned (generalisable) \\ 
    GNT\,\cite{t2023is} & \xmark & \xmark &  Learned (generalisable) \\ 
    FreeNeRF~\cite{yang2023} & \cmark & \xmark & {Frequency regularization, near penalty} \\ 
    SPARF~\cite{truong2023} & \cmark & \xmark & {MV consistency, depth consistency} \\ 
    DiffusioNeRF~\cite{wynn2023} & \cmark & \xmark & {Loss from a diffusion model on RGBD patches} \\ 
    GeCoNeRF~\cite{kwak2023} & \cmark & \xmark & {Depth-based pseudo-GT at feature level} \\ 
    FlipNeRF~\cite{seo2023} & \cmark & \xmark & {Flipped rays, near penalty, MV consistency} \\ 
    SparseNeRF~\cite{wang2023} & \cmark & \xmark & {Depth smoothness, mono-depth supervision} \\ 
    GNT-MOVE~\cite{cong2023} & \xmark & \xmark & Learned (generalisable) \\ 
    DaRF~\cite{song2023} & \cmark & \xmark & {Mono-depth supervision} \\ 
    CombiNeRF~\cite{bonotto2024} & \cmark & \xmark & {Depth smoothness, near penalty, Lipschitz reg.}  \\ 
    ConvGLR~\cite{tanay2024} & \xmark & \xmark & Learned (generalisable) \\ 
    pixelSplat~\cite{charatan24} & \xmark & \cmark & Learned (generalisable) \\ 
    MuRF~\cite{xu2024} & \cmark & \xmark & Learned (generalisable) \\ 
    DNGaussian~\cite{li2024} & \cmark & \xmark & {Mono-depth supervision} \\ 
    ZeroRF~\cite{shi2024} & \cmark & \xmark & {TensoRF with randomly-initialized generator} \\ 
    ReVoRF~\cite{xu2024b} & \cmark & \xmark & {Depth-based pseudo-GT, depth smoothing} \\ 
    Mi-MLP\,\cite{zhu2024vanilla} & \cmark & \xmark & {Input rerouting, background/sampling reg.} \\ 
    ReconFusion~\cite{wu2024} & \cmark & \xmark & {CLIP/PixelNeRF/image diffusion supervision} \\ 
    MVSplat~\cite{chen2024} & \xmark & \cmark & Learned (generalisable) \\ 
    latentSplat~\cite{wewer24latentsplat} & \xmark & \cmark & Learned (generalisable) \\ 
    CoherentGS~\cite{paliwal2024} & \xmark & \cmark & {Depth smoothness, MV consistency} \\ 
    FSGS~\cite{zhu2024} & \xmark & \cmark & {Gaussian unpooling from SfM point cloud} \\ 
    CoR-GS~\cite{zhang2024} & \xmark & \cmark & {Disagreement penalty across 3DGS models} \\ 
    CAT3D~\cite{gao2024cat3d} & \cmark & \xmark & {Multi-view diffusion supervision} \\ 
\end{tabularx}
\label{table:related_work}
\end{table}

In the sparse regime, novel view synthesis is highly under-constrained: a small number of observations can in principle be explained by infinitely many underlying 5D radiance fields. In recent years, a large number of methods have focused on this sparse regime; we present an overview in Table~\ref{table:related_work}.
These methods can all be interpreted as enforcing a specific scene prior (or a set of scene priors) to help regularize the problem, and they can be classified into two main categories: scene-specific and generalisable approaches. 

Generalisable approaches are well suited for the sparse regime by design, because they learn a scene prior during training by being exposed to a large number of scenes, and because they typically rely on a small number of input views (2 to 16). They can be divided into 4 main categories: methods that predict a multiplane image representation~\cite{zhou2018stereo,mildenhall2019local,flynn2019deepview}, methods that predict a radiance field representation~\cite{chibane2021stereo,yu2021,wang2021ibrnet,mvsnerf21,johari2022geonerf,xu2024}, methods that predict a representation with 3D Gaussians~\cite{charatan24,chen2024,wewer24latentsplat} and methods that directly predict rendered pixel colors~\cite{suhail2022generalizable,t2023is,cong2023,tanay2024}. For all these methods, multi-view consistency is typically enforced implicitly through epipolar constraints~\cite{chibane2021stereo,yu2021,wang2021ibrnet,suhail2022generalizable,t2023is,cong2023,charatan24,wewer24latentsplat}, or through the construction of cost volumes~\cite{mvsnerf21,johari2022geonerf,chen2024} or plane sweep volumes~\cite{zhou2018stereo,mildenhall2019local,flynn2019deepview,xu2024,tanay2024}.

Scene-specific approaches, on the other hand, are not particularly well suited for the sparse regime: vanilla NeRF~\cite{mildenhall2020} and 3DGS~\cite{kerbl2023} both produce degenerate outputs when trained with few images. In order to produce realistic renderings, these approaches require the addition of explicit scene priors. One common approach consists in regularizing training or target views in some way. For instance, some methods regularize the appearance of unseen patches with the help of an auxiliary network, such as a CLIP model~\cite{jain2021}, a normalizing flow model~\cite{niemeyer2022} or a diffusion model~\cite{wynn2023,wu2024}. Some methods regularize the geometry of training or target patches, by enforcing depth smoothness~\cite{niemeyer2022,wang2023,paliwal2024,bonotto2024,xu2024b} or multi-view consistency~\cite{truong2023,kwak2023,paliwal2024,seo2023}. Since 3D scenes typically consist of mostly empty space, it is also possible to enforce an emptiness prior~\cite{kim2022}, especially near the camera~\cite{yang2023,seo2023,zhu2024vanilla,bonotto2024}, or where uncertainty is high~\cite{zhang2024}. Another common approach is to augment the training data in some way. For instance, synthetic rays can be generated by flipping training rays~\cite{seo2023} or pseudo ground-truth images can be generated by reprojecting training images using predicted depth maps~\cite{xu2024b}. Additional training signal can be extracted from SfM pointclouds~\cite{roessle2022,kangle2022dsnerf,zhu2024} or distilled from depth maps predicted by pre-trained monocular depth estimators~\cite{wang2023,song2023,li2024}. Recently, state-of-the-art results have been obtained by generating pseudo ground-truth images with a multi-view diffusion model~\cite{gao2024cat3d}. Finally, a set of methods report significant improvements in the sparse regime through optimization or architectural changes. This includes masking the high-frequency components of the inputs~\cite{yang2023}, rerouting the inputs at different levels of the network~\cite{zhu2024vanilla}, parameterizing TensoRF with a randomly-initialized generator~\cite{shi2024} or applying Lipschitz regularization~\cite{bonotto2024}.

\subsection{Current Protocol Overview} \label{sec:curr_eval}

The DTU dataset has become a cornerstone of sparse neural rendering evaluation and benchmarking: it is the dataset that is most commonly used among all the methods listed in Table~\ref{table:related_work} \cite{yu2021, chibane2021stereo, jain2021, niemeyer2022, kim2022, kangle2022dsnerf, johari2022geonerf, suhail2022generalizable, yang2023, truong2023, wynn2023, seo2023, wang2023, tanay2024, xu2024, li2024, wu2024, chen2024, zhang2024}.  

\begin{figure}
    \centering
    \hfill
    \includegraphics[width=0.3\linewidth]{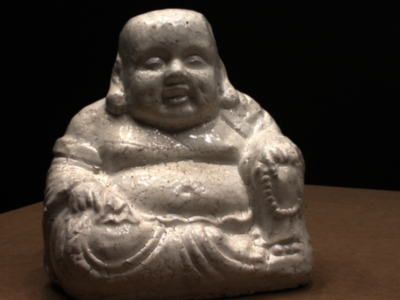}\hfill
    \includegraphics[width=0.28\linewidth]{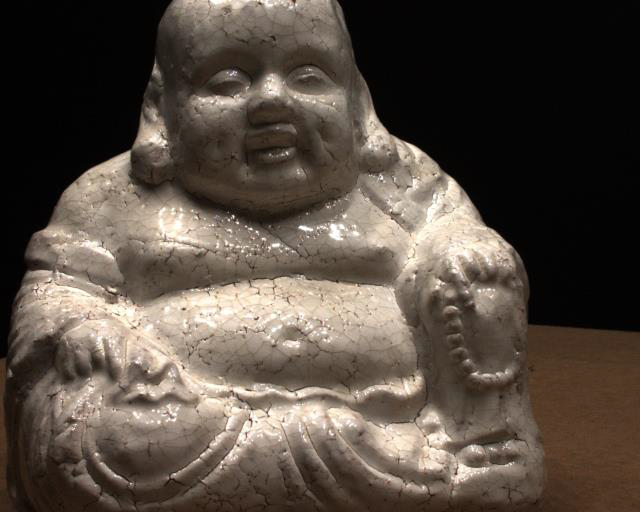}\hfill
    \includegraphics[width=0.3\linewidth]{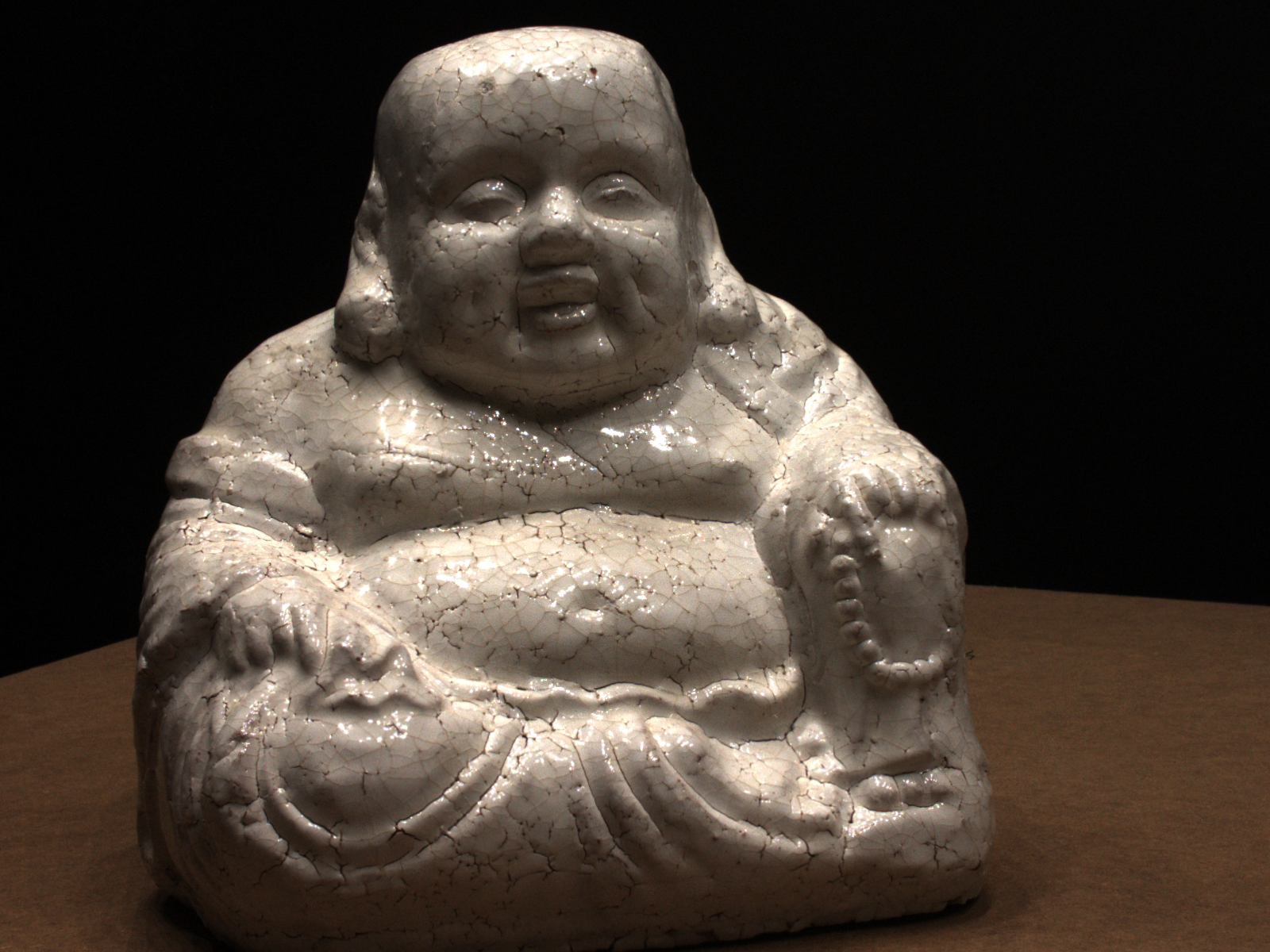}\hfill
    
    \hfill
    \makebox[0.3\textwidth]{PixelNeRF}\hfill
    \makebox[0.28\textwidth]{MVSNeRF}\hfill
    \makebox[0.3\textwidth]{Original}\hfill
    \caption{Three versions of the same image extracted from different dataset processing pipelines.} \label{fig:protocol_comparison}
\end{figure}

Unfortunately, existing protocols are somewhat outdated and prone to error. Firstly, there exists multiple versions of the dataset available online, each preprocessed in a different way. The PixelNeRF~\cite{yu2021} protocol uses a version that has been downsampled 4$\times$ using an interpolation method producing blurry images. The MVSNeRF~\cite{mvsnerf21} protocol uses a version that has been downsampled 2$\times$ and center-cropped. The RegNeRF~\cite{niemeyer2022} protocol uses the original version of the dataset and downsamples the images 4$\times$ on the fly using \textit{bilinear} interpolation (in an earlier version of the codebase) or \textit{area} interpolotation (in the current version of the codebase). 
The comparison between the images used in the evaluation is shown in Figure\,\ref{fig:protocol_comparison}.
These differences (in resolution, interpolation method and cropping), result in differences during evaluation in the metric scores, \eg PSNR, SSIM or LPIPS, which makes comparisons across methods unreliable.

To avoid these issues, we believe that the quality performance should be evaluated against the original images, hence, in full resolution and without cropping. Moreover, with recent developments in novel view synthesis, including the emergence and advances of Gaussian Splatting~\cite{kerbl2023}, the cost of training and evaluating in high resolution was drastically reduced. Furthermore, we note that the evaluation protocols used by pixelNeRF~\cite{yu2021} and RegNeRF~\cite{niemeyer2022} evaluate with the same selected training views and the same selected test views across all the objects. We believe that with the development of generalisable solutions~\cite{zhou2018stereo,yu2021,wang2021ibrnet,charatan24,chen2024,suhail2022generalizable,tanay2024}, this leads to unintended, and undetectable overfitting.

\section{Dataset}

\ds{} extends the DTU dataset \cite{aanaes2016}. We believe that a robust and fair evaluation protocol requires a larger sample of objects and images, including a subset where the test views are not publicly released. To this end, we replicate a DTU-alike setup in a synthetic Blender \cite{blender} environment. 

\subsection{Scene Composition}

\begin{figure}
    \centering
    \includegraphics[width=0.9\linewidth]{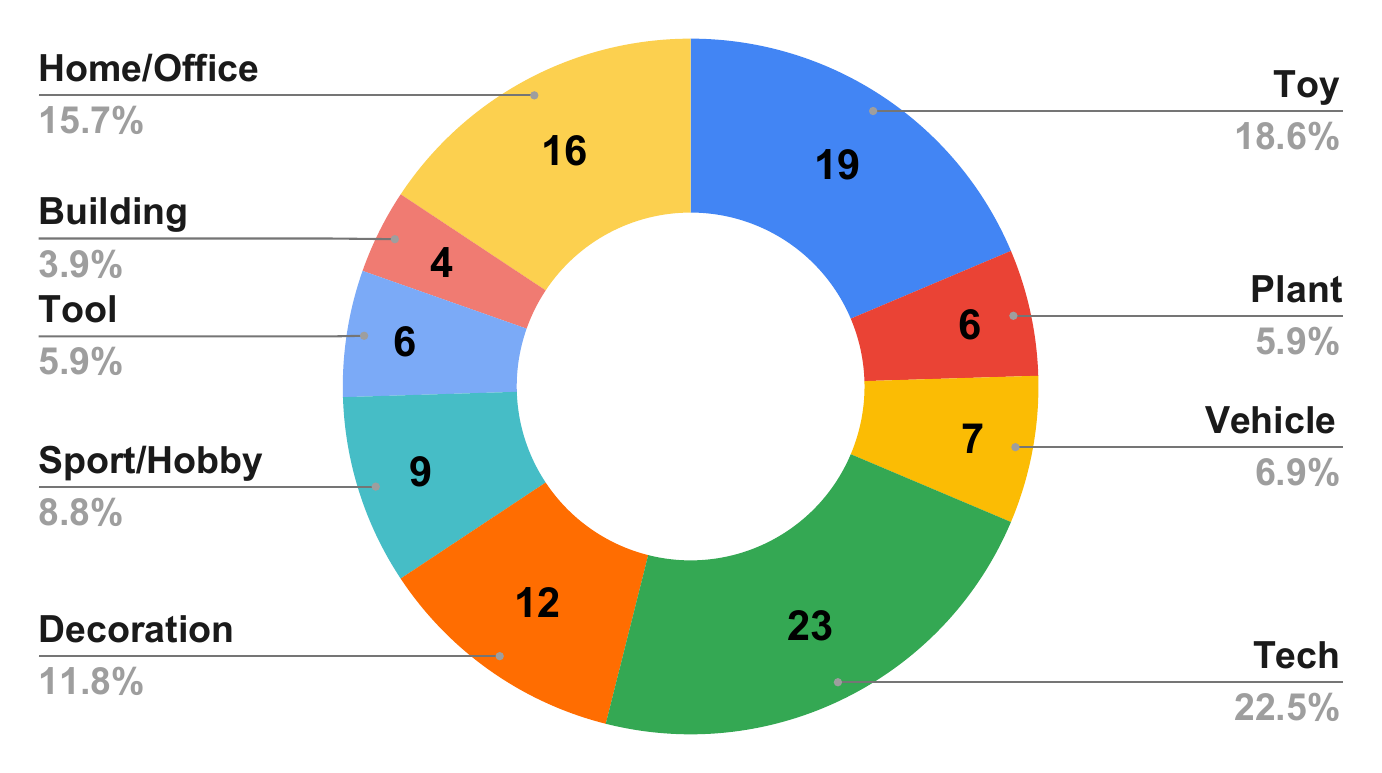}
    \caption{Composition of high-level categories of objects in \ds{}.}
    \label{fig:composition}
\end{figure}

\begin{figure}
    \centering
    \includegraphics[width=0.94\linewidth]{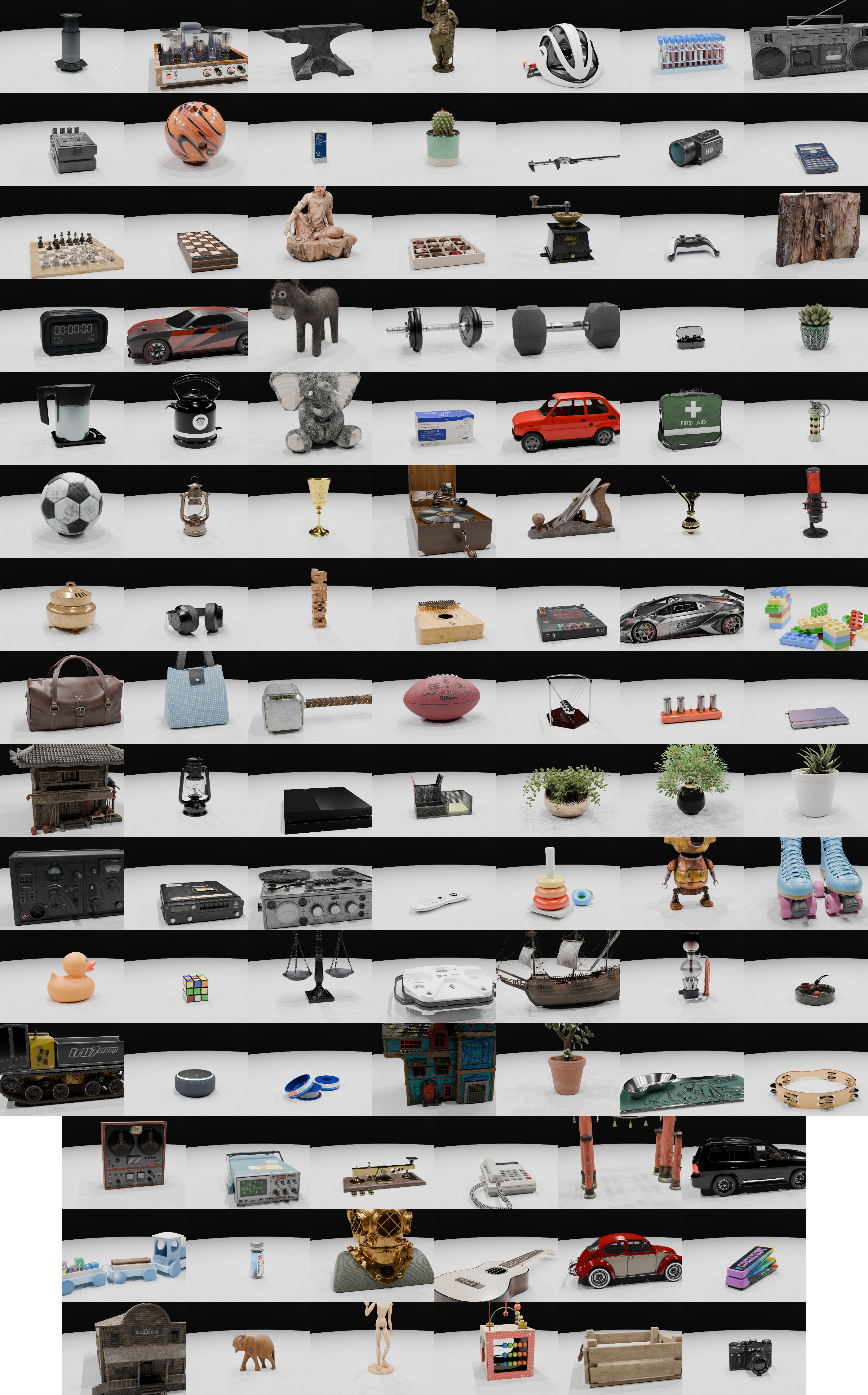}
    \caption{Examples of objects included in the \ds{} dataset. We provide objects from diverse categories with a scene placement similar to that of the DTU dataset.}
    \label{fig:examples}
\end{figure}

In an effort to replicate the setup of DTU, we place all objects on a white platform and place them in a black box. We manually collected $102$ assets from BlenderKit online library \cite{blenderkit}. All the collected models are high-quality 3D assets belonging to a range of categories, including toys, cars and houses miniatures, home appliances and equipment, technology items, tools, sports equipment, plants, decorations \etc. A breakdown of categories seen in \ds{} is presented in Figure~\ref{fig:composition}.
The scenes span a large range of objects sized between 8cm and 50cm. We include items varying in texture and specularity (\eg a plushy elephant, a glass lamp, or a shiny plastic children's toy). Selected examples from our datasets are presented in Figure \ref{fig:examples}.

\subsection{Generation}

All images in the \ds{} dataset are rendered with Blender, using a high-quality rendering engine Cycles included in Blender. We render all images in $1600\text{x}1200$ resolution. We set the virtual camera to mimic the camera used in DTU capture. To this end, we use the same focal length and the same sensor size. In our data, we eliminate the principal point shift, introduced in DTU during the process of point cloud reconstruction and image undistortion. Finally, all images are rendered together with the segmentation map used to extract object masks which are used in our evaluation protocol.

\subsection{Camera Positioning}

DTU collects the data with a setup that included a table and a robotic arm with a camera mounted on top. The camera poses in DTU are provided with respect to an arbitrary coordinate system. We exactly mimic the relative camera arrangement of DTU. However, to unify and simplify the poses, we place all the objects with the middle of their bottom side in $(0,0,0)$. Therefore, our dataset consists of $64$ camera views, $49$ placed on a smaller sphere, and $15$ on a larger, concentric sphere. Furthermore, a number of cameras in DTU capture remain unused due to a shadow cast from the robotic arm, whereas \ds{} by nature of synthetic generation does not have this limitation.

\subsection{Lighting Setup}

\begin{figure}
    \centering
    \includegraphics[width=\linewidth]{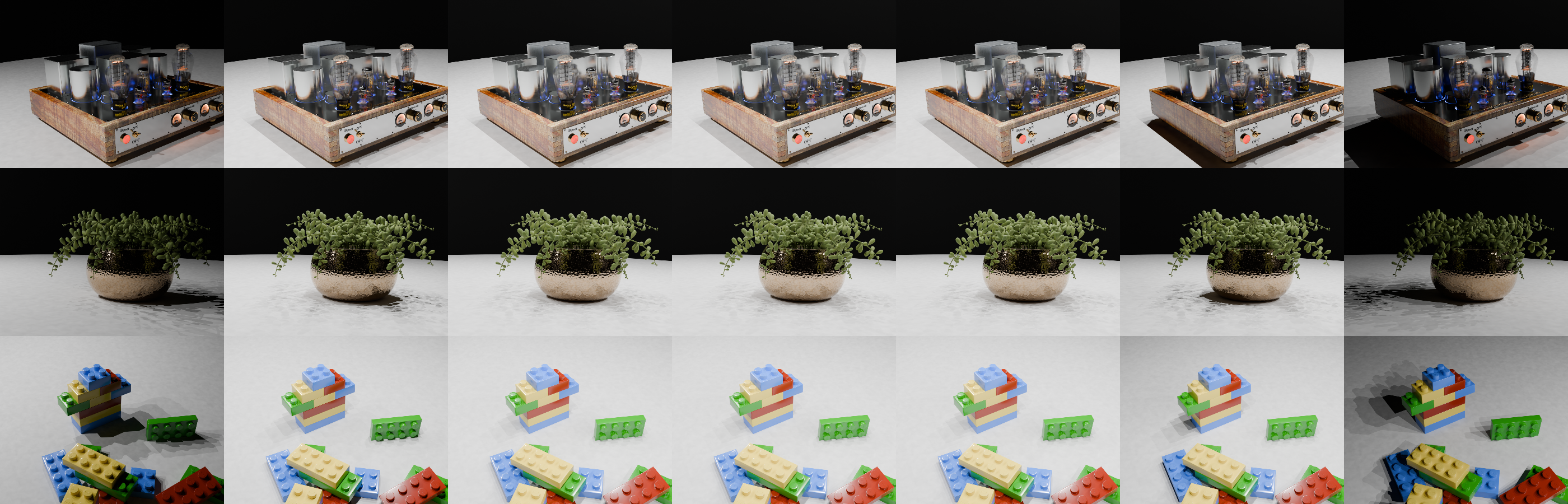}
    \caption{Varying illumination in the \ds{} dataset capture for three different scenes.}
    \label{fig:lighting}
\end{figure}

We recreate the lighting setup used by DTU. We use $16$ point lights mimicking LED sources. Following DTU, for each object we provide $7$ different captures corresponding to varying lighting conditions. This relates to the object being illuminated from either side, with one capture including all lights on. An example of
an object rendered under varying illumination is shown in Figure \ref{fig:lighting}.

\section{Evaluation Protocol}

\bench{} includes a benchmark for novel view synthesis under a sparse input image set-up. In this section, we propose and describe an evaluation protocol which we believe provides a fair comparison of sparse neural rendering methods.

\subsection{Data Splits} \label{sec:synth_split}

We propose a benchmark that includes two tracks corresponding to common setups of input views. Namely:
\begin{itemize}
    \item Track 1 - $3$ input views,
    \item Track 2 - $9$ input views.
\end{itemize}
We propose to split the \ds{} data as follows:
\begin{itemize}
    \item $82$ training scenes (for generalisable approaches),
    \item $6$ validation scenes,
    \item $9$ test scenes (including $4$ scenes common to $3$ and $9$ views benchmarks, and $5$ unique scenes each).
\end{itemize}
Note that the test split comprises some scenes shared between the two tracks and some unique to each track. This allows for comparisons between $3$ and $9$ input setups while preserving the ability to detect cross-contamination between input views in Track 1 and 2.

\subsection{Input Views} \label{sec:synth_inp_views}

\begin{figure}
    \centering
    \includegraphics[width=0.32\linewidth, clip, trim={120px 20px 60px 60px}]{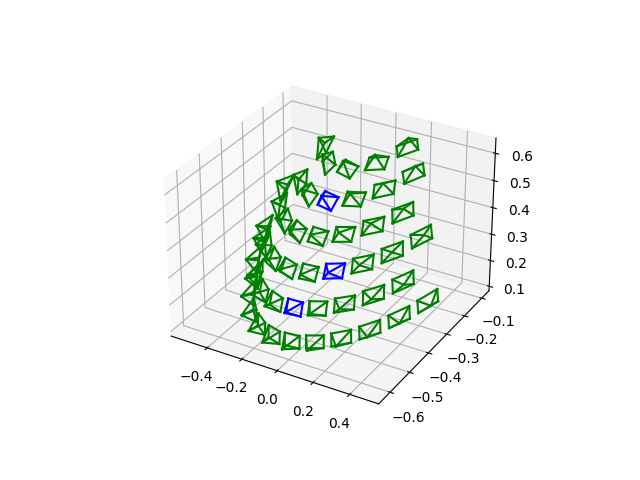}
    \includegraphics[width=0.32\linewidth, clip, trim={120px 20px 60px 60px}]{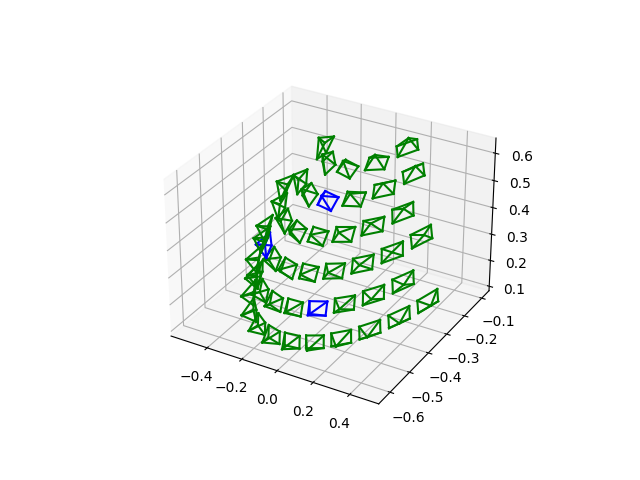}
    \includegraphics[width=0.32\linewidth, clip, trim={120px 20px 60px 60px}]{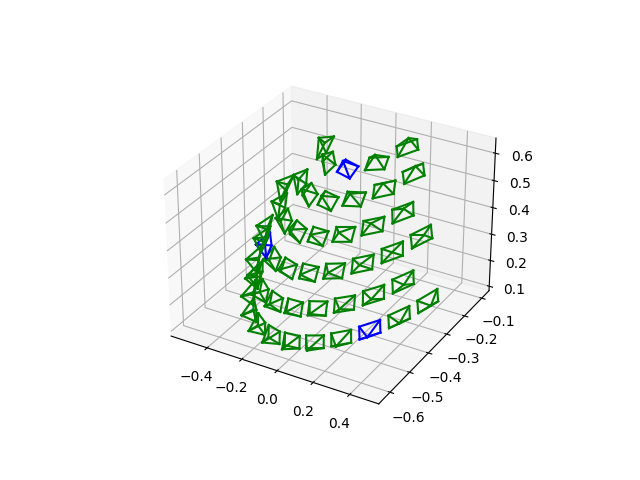}
    \caption{A diagram marking the selection of input views among $49$ available cameras. Input views are presented for the following scenes left to right: \textit{Rings}, \textit{Recorder}, \textit{Jenga}. Selected input views are marked in blue.}
    \label{fig:input_views}
\end{figure}

Motivated by the shortcomings of current uses of DTU dataset in sparse view rendering evaluations as described in Sec.\,\ref{sec:curr_eval}, we propose to evaluate \ds{} in full resolution. Further, we randomly select $3$ or $9$ camera views to serve as the input for evaluation on validation and test split. Additionally, we curate the views, especially in the $3$-view scenario to include a variety of camera distances. An example of input camera selection can be seen in Fig.\,\ref{fig:input_views}.
Similarly, we select $10$ random views from non-input camera poses for testing. We believe this allows for detailed evaluation while keeping a reasonable rendering time.
 
\subsection{Sparse Neural Rendering Challenge}

The \ds{} dataset was used in the AIM 2024 Sparse Neural Rendering Challenge. The participants were asked to develop approaches for novel view synthesis under sparse input constraints. The data used in the challenge consisted in part of the synthetic data in splits as aforementioned, and in part of DTU data. \ds{} data used in the challenge was evaluated as described in Sec.\,\ref{sec:synth_split} and Sec.\,\ref{sec:synth_inp_views}. DTU data was prepared in a similar manner to ensure a fair benchmark. To this end, we used previously proposed \cite{yu2021} $15$ evaluation scenes and used $6$ of them in the validation phase, and $9$ in the testing phase. For each of those scenes, a random selection of $3$/$9$ views was used as the inputs, and a random $10$ from the remaining ones was used as the target (having previously discarded views where the robotic arm casts a shadow onto the object). 

\subsection{Benchmark Platform}

With \bench{}, we do not release the test views publicly, instead, we provide researchers with the platform for results evaluation. The benchmark can be accessed at \url{https://sparebenchmark.github.io}.

We use Codabench\,\cite{xu2022} as the evaluation platform. For each track and split, the users can submit the rendered views which are then evaluated against the ground truth. In an effort to allow evaluation on \ds{} and preserve the AIM 2024 Sparse Neural Rendering Challenge continuity, we enable users to upload either \ds{} data results, DTU results, or exact challenge setup (both datasets) results. The benchmark automatically returns the corresponding scores. 

Given the submitted results, \bench{} returns the following metrics:
\begin{itemize}
    \item Masked PSNR (PSNR-M) - PSNR calculated with the object mask,
    \item PSNR - calculated in the whole image, including background reconstruction,
    \item SSIM-M - SSIM\,\cite{wang2004} calculated withing the tight bounding box around the object,
    \item SSIM - calculated in the whole image,
    \item LPIPS-M - LPIPS\,\cite{zhang2018} calculated withing the tight bounding box around the object, using AlexNet\,\cite{krizhevsky2012} as a backbone,
    \item LPIPS - calculated in the whole image.
\end{itemize}
The user can access a detailed report for the submission that includes the aforementioned scores for every evaluated image as well as the average for each scene.

\section{Baseline Experiments}

We perform baseline experiments in the SpaRe benchmark setting we propose. We select two methods that optimise the underlying representation for each scene separately, namely RegNeRF\,\cite{niemeyer2022} and FreeNeRF\,\cite{yang2023}. 
Further, we conduct an experiment with a generalisable method that utilises a pretaining step, \ie ConvGLR\,\cite{tanay2024}. Finally, for each Track 1 and Track 2, we include the results of the AIM 2024 Sparse Neural Rendering Challenge\,\cite{Nazarczuk2024_report} for the test set. The results of the experiment for Track 1 are presented in Table\,\ref{tab:track1}, and for Track 2 - in Table\,\ref{tab:track2}.
\begin{table}
\tablesmallsize
\setlength{\tabcolsep}{2pt}
\begin{center}
\caption{Quantitative results of baseline methods on Track 1 - 3 views on validation and test splits of SpaRe and DTU. \textsuperscript{\textdagger}winner, and \textsuperscript{\textdaggerdbl}runner-up of the AIM 2024 Sparse Neural Rendering Challenge Track 1.}
\label{tab:track1}
\begin{tabularx}{\linewidth}{Xrrrrrrrrrrrr}
\toprule
\multirow{2}{*}{Method} & \multicolumn{3}{c}{PSNR-M} & \multicolumn{3}{c}{PSNR} & \multicolumn{3}{c}{SSIM-M} & \multicolumn{3}{c}{LPIPS-M}\\
\cmidrule(lr){2-4}\cmidrule(lr){5-7}\cmidrule(lr){8-10}\cmidrule(lr){11-13}
 & Avg & DTU & Syn
& Avg & DTU & Syn
& Avg & DTU & Syn
& Avg & DTU & Syn\\
\midrule
& \multicolumn{12}{c}{Validation} \\
\cmidrule(lr){2-13}
RegNeRF\,\cite{niemeyer2022} & $18.17$ & $19.08$ & $17.27$ & $16.64$ & $14.87$ & $18.41$ & $0.591$ & $0.580$ & $0.601$ & $0.590$ & $0.587$ & $0.592$ \\
FreeNeRF\,\cite{yang2023} & $17.55$ & $18.53$ & $16.57$ & $17.17$ & $15.48$ & $18.85$ & $0.585$ & $0.568$ & $0.601$ & $0.535$ & $0.559$ & $0.510$ \\
ConvGLR\,\cite{tanay2024} & $20.81$ & $21.86$ & $19.75$ & $23.32$ & $21.80$ & $24.83$ & $0.672$ & $0.667$ & $0.676$ & $0.549$ & $0.564$ & $0.534$ \\
\midrule
& \multicolumn{12}{c}{Test} \\
\cmidrule(lr){2-13}
RegNeRF\,\cite{niemeyer2022} & $17.47$ & $17.76$ & $17.17$ & $16.24$ & $15.84$ & $16.64$ & $0.629$ & $0.553$ & $0.704$ & $0.533$ & $0.562$ & $0.504$ \\
FreeNeRF\,\cite{yang2023} & $17.93$ & $16.61$ & $19.26$ & $17.62$ & $15.93$ & $19.30$ & $0.633$ & $0.542$ & $0.723$ & $0.462$ & $0.553$ & $0.371$ \\
ConvGLR\,\cite{tanay2024} & $20.11$ & $20.36$ & $19.87$ & $22.60$ & $20.87$ & $24.33$ & $0.710$ & $0.644$ & $0.776$ & $0.469$ & $0.534$ & $0.405$ \\
FrameNeRF\textsuperscript{\textdagger}\!\!\,\cite{xing2024} & $18.67$ & $18.50$ & $18.83$ & $17.98$ & $16.73$ & $19.23$ & $0.665$ & $0.591$ & $0.740$ & $0.395$ & $0.420$ & $0.369$ \\
SCNeRF\textsuperscript{\textdaggerdbl}\!\!\,\cite{li2024_scnerf} & $18.30$ & $18.16$ & $18.43$ & $18.18$ & $17.00$ & $19.36$ & $0.654$ & $0.584$ & $0.725$ & $0.515$ & $0.584$ & $0.447$ \\
\bottomrule
\end{tabularx}
\end{center}
\end{table}

\begin{table}
\tablesmallsize
\setlength{\tabcolsep}{2pt}
\begin{center}
\caption{Quantitative results of baseline methods on Track 2 - 9 views on validation and test splits of SpaRe and DTU. \textsuperscript{\textdagger}winner, and \textsuperscript{\textdaggerdbl}runner-up of AIM 2024 Sparse Neural Rendering Challenge - Track 2.}
\label{tab:track2}
\begin{tabularx}{\linewidth}{Xrrrrrrrrrrrr}
\toprule
\multirow{2}{*}{Method} & \multicolumn{3}{c}{PSNR-M} & \multicolumn{3}{c}{PSNR} & \multicolumn{3}{c}{SSIM-M} & \multicolumn{3}{c}{LPIPS-M}\\
\cmidrule(lr){2-4}\cmidrule(lr){5-7}\cmidrule(lr){8-10}\cmidrule(lr){11-13}
 & Avg & DTU & Syn
& Avg & DTU & Syn
& Avg & DTU & Syn
& Avg & DTU & Syn\\
\midrule
& \multicolumn{12}{c}{Validation} \\
\cmidrule(lr){2-13}
RegNeRF\,\cite{niemeyer2022} & $23.74$ & $25.50$ & $21.98$ & $22.51$ & $22.16$ & $22.87$ & $0.697$ & $0.694$ & $0.701$ & $0.477$ & $0.504$ & $0.450$ \\
FreeNeRF\,\cite{yang2023} & $25.15$ & $27.03$ & $23.26$ & $25.15$ & $25.48$ & $24.82$ & $0.746$ & $0.736$ & $0.757$ & $0.328$ & $0.386$ & $0.271$ \\
ConvGLR\,\cite{tanay2024} & $23.42$ & $25.33$ & $21.51$ & $26.61$ & $25.81$ & $27.40$ & $0.712$ & $0.709$ & $0.715$ & $0.479$ & $0.505$ & $0.453$ \\
\midrule
& \multicolumn{12}{c}{Test} \\
\cmidrule(lr){2-13}
RegNeRF\,\cite{niemeyer2022} & $23.22$ & $23.24$ & $23.21$ & $21.29$ & $20.95$ & $21.62$ & $0.723$ & $0.688$ & $0.758$ & $0.415$ & $0.451$ & $0.379$ \\
FreeNeRF\,\cite{yang2023} & $24.27$ & $24.24$ & $24.30$ & $23.51$ & $23.20$ & $23.83$ & $0.759$ & $0.731$ & $0.786$ & $0.293$ & $0.336$ & $0.251$ \\
ConvGLR\,\cite{tanay2024} & $22.62$ & $22.87$ & $22.37$ & $24.59$ & $23.80$ & $25.37$ & $0.731$ & $0.697$ & $0.765$ & $0.419$ & $0.459$ & $0.379$ \\
FrameNeRF\textsuperscript{\textdagger}\!\!\,\cite{xing2024} & $24.51$ & $24.56$ & $24.46$ & $23.87$ & $23.79$ & $23.94$ & $0.784$ & $0.759$ & $0.808$ & $0.262$ & $0.267$ & $0.257$ \\
Thirteen\textsuperscript{\textdaggerdbl} & $21.59$ & $20.14$ & $23.04$ & $21.45$ & $19.73$ & $23.16$ & $0.649$ & $0.549$ & $0.749$ & $0.516$ & $0.628$ & $0.403$ \\
\bottomrule
\end{tabularx}
\end{center}
\end{table}

In the quantitative comparison, we report the results as masked PSNR (PSNR-M), PSNR calculated in the whole image, and SSIM and LPIPS calculated within a tight bounding box around the object (SSIM-M and LPIPS-M).

\subsection{Track 1}

\begin{figure}
    \centering
    \includegraphics[width=0.6\linewidth]{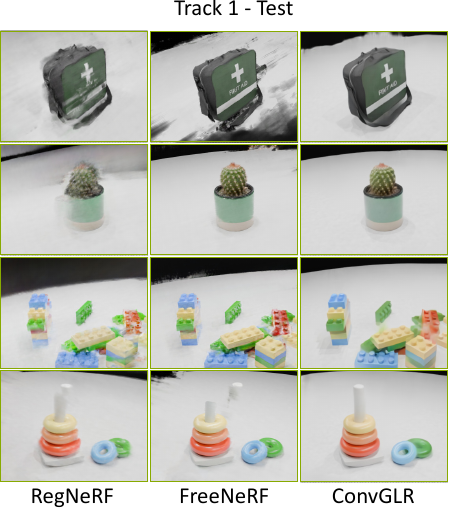}
    \caption{Test set results on the SpaRe synthetic dataset for Track 1. Ground truth images are omitted to preserve benchmark integrity.}
    \label{fig:results-synthetic-t1}
\end{figure}

\begin{figure}
    \centering
    \includegraphics[width=0.8\linewidth]{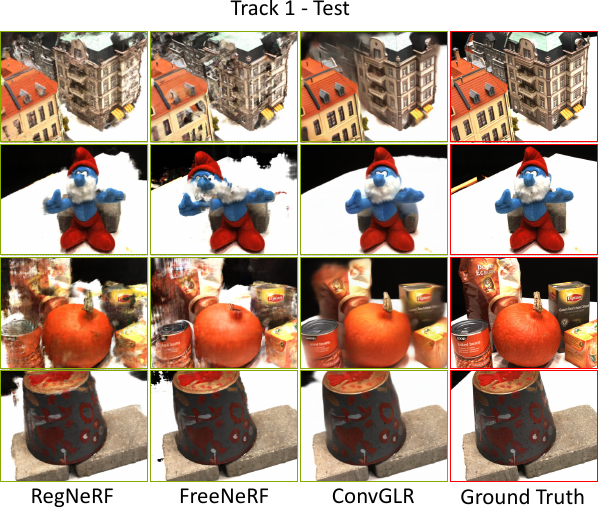}
    \caption{Test set results on the DTU dataset for Track 1.}
    \label{fig:results-dtu-t1}
\end{figure}

The quantitative results for Track 1 (3 input views), for both validation and test splits of SpaRe benchmark (including SpaRe and DTU datasets) are presented in Table\,\ref{tab:track1}. Figure\,\ref{fig:results-synthetic-t1} shows the visualisation of the baseline models evaluated on the SpaRe synthetic dataset. Note that ground truth images are withheld to preserve the integrity of the benchmark (the evaluation platform with detailed results is publicly available). Similarly, Figure\,\ref{fig:results-dtu-t1} presents visualisations of novel views on DTU data in the setting proposed by SpaRe benchmark.  

We observe the superior performance of ConvGLR in a 3 input views scenario across the majority of provided metrics (PSNR-M, PSNR, SSIM-M). Given a single scene, the task of reconstructing 3D representation is highly underconstrained. We believe that in the very sparse scenario, the method pretrained on a large dataset is able to learn stronger geometry priors than models that optimise the representation on a per-scene basis. Notably, we observe the largest performance gap in PSNR calculated on the whole image, namely, $4.42dB$ higher than the runner-up. This can be attributed to ConvGLR having seen the background throughout the training data, and being able to reconstruct its content in the test set. Interestingly, we see that the winner of the AIM 2024 Sparse Neural Rendering Challenge performing better in one of the perceptual metrics (LPIPS-M) than ConvGLR. In the qualitative examples in Figure\,\ref{fig:results-synthetic-t1} we observe corresponding relations. Across all the images, we can see that ConvGLR is the only method that reconstructs both the object and the background without artefacts. In contrast, per-scene optimisation methods include artefacts as a result of 3 views not constraining the geometry enough, \eg see the tip of the Hanoi Tower toy. We observe a similar behaviour on the DTU dataset shown in Figure\,\ref{fig:results-dtu-t1}. Note Papa Smurf's head artefact for FreeNeRF, or blurred object silhouettes in the grocery scene. On the other hand, we observed a good performance of FreeNeRF in perceptual metric - LPIPS-M. This can be observed in \eg the sharp texture of the cactus from the SpaRe dataset (Fig.\,\ref{fig:results-synthetic-t1}). 

In summary, we observe that in the scenario where only 3 input views are available, the generalisable method provides very strong object priors capable of regularising the underlying geometry, whereas per-scene optimisation may recover some higher frequency details suffering, however, from artefact induced by sparse supervision.

\subsection{Track 2}

\begin{figure}
    \centering
    \includegraphics[width=0.6\linewidth]{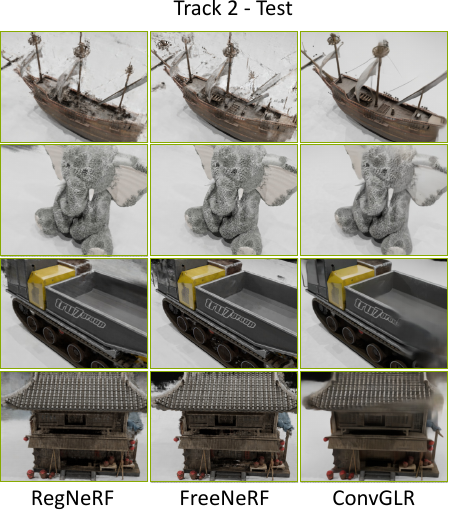}
    \caption{Test set results on the synthetic dataset for Track 2. Ground truth images are omitted to preserve benchmark integrity.}
    \label{fig:results-synthetic-t2}
\end{figure}

\begin{figure}
    \centering
    \includegraphics[width=0.8\linewidth]{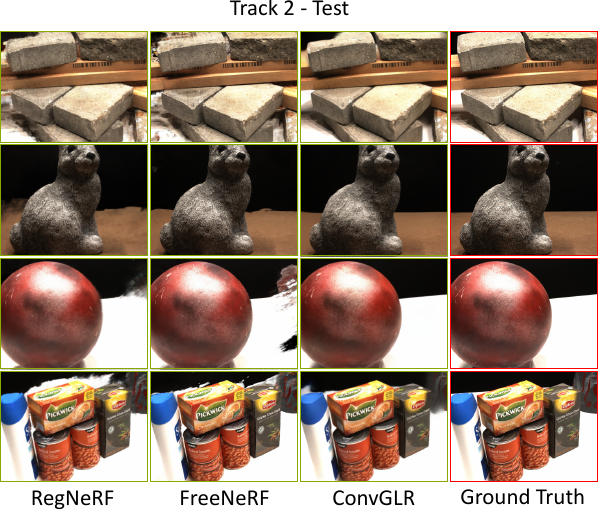}
    \caption{Test set results on the DTU dataset for Track 2.}
    \label{fig:results-dtu-t2}
\end{figure}

The quantitative results for Track 2 (9 input views), for both validation and test splits of SpaRe benchmark (including SpaRe and DTU datasets) are presented in Table\,\ref{tab:track2}. Figure\,\ref{fig:results-synthetic-t2} shows the visualisation of the baseline models evaluated on the SpaRe synthetic dataset.
Figure\,\ref{fig:results-dtu-t2} presents visualisations of novel views on DTU data in the setting proposed by SpaRe benchmark.  

In Track 2, we observe results contrasting to those of Track 1. We notice that the methods optimising the representation separately for each scene perform better than the generalisable one (see PSNR-M for FrameNeRF and FreeNeRF $24.51dB$ and $24.27dB$ in contrast to ConvGLR $22.62dB$). This leads us to believe that current per-scene optimisation methods can utilise priors from 9 input views to a better extent than a generalisable prior obtained in pertaining. However, we still observe a performance gap in favour of ConvGLR when PSNR is calculated over the whole image. This affirms the belief that generalisable methods having seen significantly more examples of the background can reconstruct it better. In Figure\,\ref{fig:results-synthetic-t2} we present qualitative results of the Track 2 test set from SpaRe data. We observe similarities to the corresponding results in Track 1. RegNeRF and FreeNeRF seem to provide sharper high-frequency details, \eg plush texture of the elephant, the writing on the snow truck, or the wood boards on the house model. Similarly, ConvGLR provides smoother results around the edges and exhibits fewer artefacts. However, in Track 2 the artefacts of per-scene optimisation models are much smaller than in Track 1 which is reflected in high scores in performance metrics for these algorithms. A similar visual comparison of novel views in DTU dataset is shown in Figure\,\ref{fig:results-dtu-t2}. Here, the observations are similar to SpaRe dataset. We notice smoother reconstruction around the edges for ConvGLR, whereas FreeNeRF and RegNeRF seem to provide sharper high-frequency details, \eg barcode on the wooden board, or bunny statue texture.

\subsection{Ablation Study}

In this section we experiment with different training data setups used within the generalisable method ConvGLR. In Table\,\ref{tab:track2_transferability} we show the results of training ConvGLR on SpaRe training set, DTU training set, and mix of all data. We perform experiments consistent with Track 2 of our benchmark \ie 9 input views scenario.

\begin{table}
\tablesmallsize
\setlength{\tabcolsep}{2pt}
\begin{center}
\caption{Quantitative results of generalisable method -  ConvGLR on Track 2 when trained with different data.}
\label{tab:track2_transferability}
\begin{tabularx}{\linewidth}{Xrrrrrrrrrrrr}
\toprule
\multirow{2}{*}{Training data} & \multicolumn{3}{c}{PSNR-M} & \multicolumn{3}{c}{PSNR} & \multicolumn{3}{c}{SSIM-M} & \multicolumn{3}{c}{LPIPS-M}\\
\cmidrule(lr){2-4}\cmidrule(lr){5-7}\cmidrule(lr){8-10}\cmidrule(lr){11-13}
 & Avg & DTU & Syn
& Avg & DTU & Syn
& Avg & DTU & Syn
& Avg & DTU & Syn\\
\midrule
& \multicolumn{12}{c}{Validation} \\
\cmidrule(lr){2-13}
DTU & $22.76$ & $25.38$ & $20.14$ & $22.18$ & $26.22$ & $18.14$ & $0.709$ & $0.721$ & $0.697$ & $0.488$ & $0.482$ & $0.494$ \\
Synthetic & $20.76$ & $20.24$ & $21.29$ & $22.11$ & $16.63$ & $27.60$ & $0.665$ & $0.618$ & $0.711$ & $0.531$ & $0.602$ & $0.460$ \\
DTU + Synth & $23.42$ & $25.33$ & $21.51$ & $26.61$ & $25.81$ & $27.40$ & $0.712$ & $0.709$ & $0.715$ & $0.479$ & $0.505$ & $0.453$ \\
\midrule
& \multicolumn{12}{c}{Test} \\
\cmidrule(lr){2-13}
DTU & $21.80$ & $23.17$ & $20.42$ & $21.15$ & $24.29$ & $18.00$ & $0.723$ & $0.712$ & $0.735$ & $0.437$ & $0.440$ & $0.434$ \\
Synthetic & $20.33$ & $18.51$ & $22.16$ & $20.82$ & $15.95$ & $25.69$ & $0.684$ & $0.606$ & $0.763$ & $0.475$ & $0.563$ & $0.386$ \\
DTU + Synth & $22.62$ & $22.87$ & $22.37$ & $24.59$ & $23.80$ & $25.37$ & $0.731$ & $0.697$ & $0.765$ & $0.419$ & $0.459$ & $0.379$ \\
\bottomrule
\end{tabularx}
\end{center}
\end{table}

We observe that both datasets present a similar performance in generalisation capability, performing the best on the data trained with the corresponding dataset. We can see that the best average performance is achieved when a mix of training data is used. Notably, we observe a high impact of the background on the performance of the model \eg there is a large gap in PSNR calculated on the whole image between datasets when the model has seen only one. This indicates that the inclusion of higher-variety data is beneficial for the generalisable approaches.

\section{Conclusions}

This paper introduces a new benchmark and dataset for Sparse Neural Rendering - SpaRe. In this work, we identify shortcomings of current benchmark protocols based on the DTU dataset. This includes but is not limited to low-resolution evaluation, diverse pre-processing step, and using the same camera position as testing views across all the scenes. To alleviate these issues, we propose a new dataset, SpaRe, composed by high-quality renderings in a setup that reproduces that of the DTU capture. Further, we suggest a new benchmarking protocol for the SpaRe dataset and DTU dataset, that introduces more variety in the poses of the input camera views. To facilitate a unified and fair evaluation, we propose to keep the ground-truth images of target scenes and views secret. To this end, we provide an online platform for submitting the results and their evaluation. This tool fosters reproducible evaluation, and enables
researchers easy access to a public leaderboard with the state-of-the-art
performance scores that can be updated on a rolling basis. In the experimental section, we investigate several state-of-the-art baselines, emphasising the advantages and drawbacks of per-scene optimisation, and generalisable approaches. When training generalisable methods, using SpaRe and the DTU dataset yields the best possible results for both synthetic and real data.

\section*{Acknowledgements}
This work was partially supported by the Humboldt Foundation. We thank the AIM 2024 sponsors: Meta Reality Labs, KuaiShou, Huawei, Sony Interactive Entertainment and University of W\"urzburg (Computer Vision Lab).

\bibliographystyle{splncs04}
\bibliography{main}
\end{document}